\documentclass{applemlr}
\usepackage{amsmath}
\usepackage{enumerate} 
\usepackage{algorithm}
\usepackage{algpseudocode}
\usepackage{amsfonts}
\usepackage{amsthm}
\usepackage{newtxtt}
\usepackage{cleveref}
\usepackage{diagbox}
\usepackage{colortbl}
\usepackage{amssymb}
\usepackage{xspace}
\usepackage{wrapfig}
\usepackage{adjustbox}
\usepackage{tabularx}
\usepackage{booktabs}
\usepackage{mathtools}
\usepackage{tikz}
\usepackage{enumitem}
\usepackage{silence}
\usepackage{dsfont}
\usepackage[table]{xcolor}
\usepackage[dvipsnames]{xcolor}
\usepackage{multirow}
\usepackage{makecell}
\usepackage{arydshln}
\usepackage{xfakebold}

\usepackage{amsmath,amsfonts,bm}

\def\eqref#1{equation~\ref{#1}}

\def\1{\bm{1}}

\DeclareMathAlphabet{\mathsfit}{\encodingdefault}{\sfdefault}{m}{sl}
\SetMathAlphabet{\mathsfit}{bold}{\encodingdefault}{\sfdefault}{bx}{n}

\definecolor{textgray}{HTML}{6E6E73}
\usetikzlibrary{positioning, calc}
\usetikzlibrary{decorations.pathmorphing}

\makeatletter
\patchcmd{\wrong@fontshape}{\@gobbletwo}{}{}{}
\makeatother
\numberwithin{equation}{section} 
\tcbuselibrary{minted}
\setminted[python]{
  linenos,
  breaklines,
  fontsize=\footnotesize,
  xleftmargin=2em
}

\makeatletter
\AtBeginDocument{
  \urlstyle{sf}
  
}
\makeatother

\definecolor{light}{RGB}{125, 125, 125}
\crefname{tcb@cnt@pbox}{code}{code}
\Crefname{tcb@cnt@pbox}{Code}{Code}
\crefname{assumption}{assumption}{assumption}
\Crefname{assumption}{Assumption}{Assumptions}

\newtcolorbox[auto counter]{pbox}[2][]{
  colback=white,
  title=Code~\thetcbcounter: #2,
  #1,fonttitle=\sffamily,
  fontupper=\sffamily,
  arc=2pt,
  colframe=bgcolor,
  coltitle=fgcolor,
  colbacktitle=bgcolor,
  toptitle=0.25cm,
  bottomtitle=0.125cm
}

\makeatletter
\newcommand\applefootnote[1]{%
  \begingroup
  \renewcommand\thefootnote{}%
  \renewcommand\@makefntext[1]{\noindent##1}%
  \footnote{#1}%
  \addtocounter{footnote}{-1}%
  \endgroup
}
\makeatother

\definecolor{cverbbg}{gray}{0.90}

\newcommand{\pathmoe}{\texttt{PathMoE}}

\title{Path-Constrained Mixture-of-Experts}

\author{Zijin Gu}
\author{Tatiana Likhomanenko}
\author{Vimal Thilak}
\author[\dag]{Jason Ramapuram}
\author[\dag]{Navdeep Jaitly}

\affiliation{Apple}
\affiliation[\dag]{Google}
\contribution[\dag]{Work done at Apple.}

\abstract{
Sparse Mixture-of-Experts (MoE) architectures route each token through a subset of experts at each layer independently.
We propose viewing MoE computation through the lens of \emph{expert paths}---the sequence of expert selections a token makes across all layers.
This perspective reveals that, despite $N^L$ possible paths for $N$ experts
across $L$ layers, tokens in practice cluster into a small fraction of paths
that align with linguistic function, yet the vast majority of paths remain unexplored, representing a statistical inefficiency.
This motivates architectures that constrain the effective path space to amplify this natural concentration.
As one instantiation, we introduce \pathmoe{}, which shares router parameters
across blocks of consecutive layers.
Analysis confirms that \pathmoe{} amplifies the emergent path structure: it produces more concentrated path clusters, better cross-layer consistency, and greater robustness to routing perturbations.
Experiments on 0.9B and 16B parameter \pathmoe{} models demonstrate consistent improvements on perplexity and downstream tasks over independent routing, while eliminating the need for auxiliary losses.
These results establish expert paths as a useful design axis for MoE
architectures, complementary to existing work on independent routing mechanisms.}

\metadata[Correspondence]{\sffamily Zijin Gu: \url{zijin@apple.com}}
\date{\sffamily\today}

\begin{document}

\maketitle

\section{Introduction}
\begin{figure}[h]
  \centering
  \includegraphics[width=0.6\columnwidth]{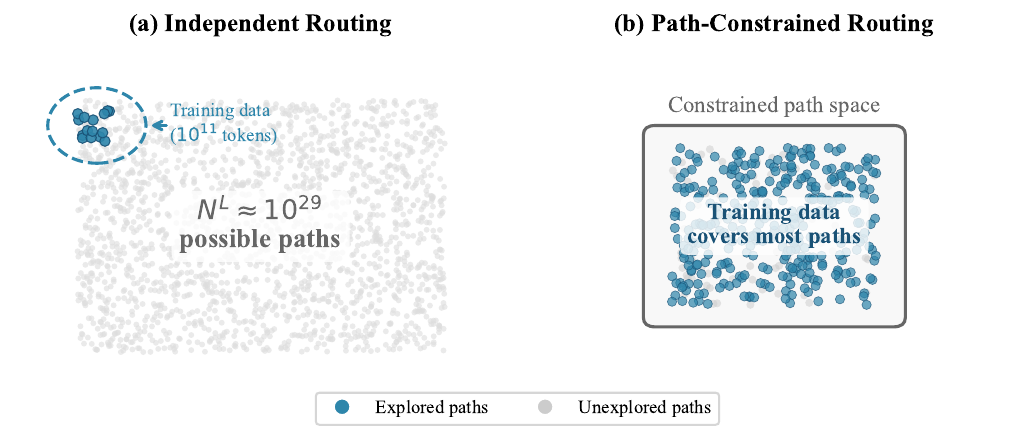}
  \caption{Statistical inefficiency of independent routing.}
  \label{fig:statistical_motivation}
\end{figure}

Scaling drives progress in deep learning, but dense models incur computational costs that grow linearly with parameter count.
Mixture-of-Experts (MoE) architectures address this by activating only a subset
of parameters for each input, decoupling model capacity from computational cost
and enabling models with trillions of parameters within practical
budgets~\citep{fedus2022switch, jiang2024mixtral}.

Most work on MoE has focused on improving \emph{individual} routing decisions,
e.g., how each layer selects experts for a given token.
In this paper, we propose a complementary perspective: viewing MoE computation through the lens of \emph{expert paths}---the full sequence of expert selections $(e_1, e_2, \ldots, e_L)$ a token makes across $L$ layers.

This path perspective reveals an important observation.
With $N$ experts per layer and independent routing, $N^L$ possible paths exist, far exceeding typical training set sizes (Figure \ref{fig:statistical_motivation}).~\footnote{
Concretely, a model with 24 layers and 16 experts per layer has $\approx
10^{29}$ possible paths---orders of magnitude larger than the number of tokens
reported in typical large model training runs ($\sim
10^{13}$).}
Yet MoE models train well in practice. Analyzing trained models, we find that
\textit{tokens naturally cluster into a small fraction of paths, with different
paths specializing for different linguistic functions} (Section~\ref{sec:path_analysis}).
While this emergent concentration of paths mitigates the combinatorial challenge, the vast majority of paths remain unexplored, representing a statistical inefficiency.

\begin{figure}[t]
  \centering
  \includegraphics[width=\textwidth]{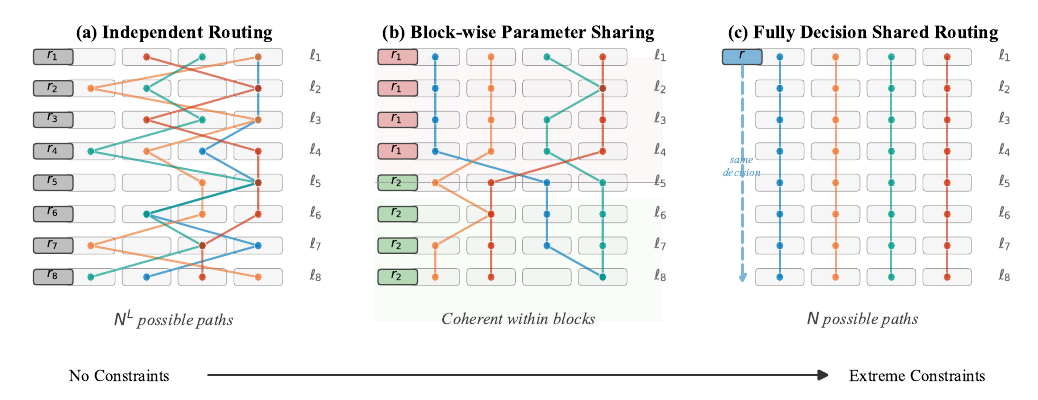}
  \caption{Spectrum of routing constraints in MoE architectures.
  \textbf{(a)} Independent routing:each layer has its own router $r_i$, creating $N^L$ possible paths for $N$ experts and $L$ layers. 
  \textbf{(b)} Block-wise parameter sharing routing, dubbed \pathmoe{}: layers within a block share router parameters.
  \textbf{(c)} Fully decision shared routing: all layers share one router and its decision.}
  \label{fig:motivation}
\end{figure}

This raises a natural question: \textbf{can we design architectures that constrain the path space to amplify the concentration?}
Figure~\ref{fig:motivation} illustrates a spectrum of possible constraints: from fully independent routing ($N^L$ paths) through block-wise sharing to fully shared routing decisions ($N$ paths).
As a concrete instantiation, we explore \emph{sharing router parameters across blocks of consecutive layers}, dubbed \pathmoe{} (Figure \ref{fig:motivation}(b)).
This creates an inductive bias where the same routing function is applied to gradually-evolving token representations within each block.
Since nearby layers process similar representations (thanks to residual
connections), they naturally receive similar (though not identical) routing,
constraining the path space without collapsing it, while allowing different
blocks to adapt to the changing nature of representations.

Using \pathmoe{} as a test bed for the path-constrained perspective, we find key benefits over conventional routing:
\begin{enumerate}
    \item \textbf{More concentrated expert paths.} \pathmoe{} amplifies the natural path clustering, producing more concentrated path distributions with 11\% lower routing entropy.
    \item \textbf{Consistent performance gains.} Experiments across 0.9B and 16B parameter models show both improved accuracy on downstream tasks and lower perplexity.
    \item \textbf{No auxiliary load balancing loss.} \pathmoe{} maintains balanced expert utilization without auxiliary losses, eliminating an extra hyperparameter.
    \item \textbf{Improved cross-layer coordination and robustness.} \pathmoe{} achieves 31\% higher routing consistency and 22.5$\times$ greater robustness to routing perturbations.
\end{enumerate}

These results establish \emph{expert paths} as a useful design axis for MoE
architectures: how tokens are distributed across cross-layer expert combinations
can be directly shaped through architectural
choices, complementing existing work on per-layer routing mechanisms. \pathmoe{}
is one simple instantiation of this principle; we expect the path perspective to
motivate further designs beyond parameter sharing.

\section{Preliminaries}\label{sec:prelim}
\vspace{-0.3cm}
\paragraph{MoE Routing.} In a transformer model with MoE layers, each MoE layer contains $N$ expert networks $\{F_1, \ldots, F_N\}$ and a router $r$. Given input token representation $\mathbf{x}$ for an MoE layer, the router computes expert probabilities~$\mathbf{p}$:
\begin{equation}
    \mathbf{p} = \mathrm{softmax}[r(\mathbf{x})] = \mathrm{softmax}\left[\mathbf{W} \mathbf{x}\right], \,\,\, \mathbf{p}\in \mathbb{R}^N.
\end{equation}
Then the output $\mathbf{y}$ is a weighted sum over the top-$k$ experts:
\begin{equation}
    \mathbf{y} = \sum_{i \in \mathrm{TopK}(\mathbf{p})} p_i \cdot F_i(\mathbf{x}).
\end{equation}
Let us consider an MoE network with $L$~layers and $N$~experts per layer.
In standard MoE, each layer~$l$ has its own learnable router parameters $\mathbf{W}_l$; we call this \textit{independent routing}.

\vspace{-0.3cm}
\paragraph{Auxiliary Load Balancing Loss.}
To prevent expert collapse where only a few experts receive most tokens, MoE training typically adds an auxiliary loss~\citep{fedus2022switch}:
\begin{equation}
    \mathcal{L}_{\text{aux}} = \alpha \cdot N \cdot \sum_{i=1}^{N} f_i \cdot P_i,
\end{equation}
where $f_i$ is the fraction of tokens routed to expert $i$, $P_i$ is the average routing probability for expert $i$, and $\alpha$ is a weighting coefficient.
This loss encourages uniform expert utilization but introduces a hyperparameter
that needs tuning.

\vspace{-0.3cm}
\paragraph{Expert Path.}
Let $[N]=\{1,\dots,N\}$, data follow distribution $\mathbf{x}\sim \mathcal{D}$, and $E_l$ be a random variable of the expert selection at layer~$l$.
Denote expert selections across $L$~layers as $\mathbf{E} = (E_1, \ldots, E_L)$, called an \textit{expert path}, which is a random variable taking values in $[N]^L$.
Each $E_l$ follows a categorical distribution over $N$ experts: $E_l \mid \mathbf{x}_l \sim \text{Categorical}\left[ \mathrm{softmax}(\mathbf{W}_l \mathbf{x}_l) \right]$, where for independent routing all $\mathbf{W}_l$ are independent matrices.
Let $\mathbf{e} = (e_1, \dots, e_L)$ be a specific realization of $\mathbf{E}$.
We consider top-1 expert for the purpose of path analysis~\footnote{Top-1 selection yields a single discrete path per token, making path statistics well-defined. Our method uses top-$k$ during training.}.

\vspace{-0.3cm}
\paragraph{Routing Entropy.}\label{def:routing-entropy}
To quantify the \textit{effective size of the expert path space}, we define the routing entropy.
Let $P_\pi(\mathbf{e})$ denote the marginal probability of observing path $\mathbf{e}$ over the data distribution $\mathcal{D}$:
$P_\pi(\mathbf{e}) =
\mathbb{E}_{\mathbf{x} \sim \mathcal{D}} \left[P(E_1=e_1,\ldots, E_L=e_L \mid \mathbf{x})\right]$.
The \emph{routing entropy} is the Shannon entropy on this marginal distribution of expert paths over the data distribution $\mathcal{D}$:
\begin{equation}
    H(\mathbf{E}) = H(E_1, \ldots, E_L) = -\sum_{\mathbf{e} \in [N]^L} P_\pi(\mathbf{e}) \log P_\pi(\mathbf{e}).
\end{equation}
A higher H(E) indicates that the model utilizes a more diverse set of expert combinations across the data.
Computationally, the marginal distribution is intractable, thus in practice we
compute an empirical routing entropy on finite data (see Appendix~\ref{app:theory_validation} for details).

\section{\pathmoe{}}\label{sec:theory}
\subsection{\pathmoe{}: Block-wise Parameter Shared Routing}

\paragraph{\pathmoe{} Routing.}
\pathmoe{} shares router parameters across blocks of consecutive layers. Given $L$ MoE layers and block size $B$, we partition layers into $\lceil L/B \rceil$ blocks. All layers within a block share a single router:
\begin{equation}
    r_l = r^{(\lceil l/B \rceil)}_{\text{shared}}, \quad \forall l \in \{1, \ldots, L\}.
\end{equation}

\paragraph{Expert Path for \pathmoe{}.}
For \pathmoe{} routing, each $E_l$ still follows a categorical distribution over $N$ experts.
However, for any layer $l'$ in the same block $b=\lceil l'/B \rceil$ we have $E_{l'} \mid \mathbf{x}_{l'} \sim \text{Categorical}\left[ \mathrm{softmax}(\mathbf{W}_b \mathbf{x}_{l'}) \right]$, where $\mathbf{W}_b$ from different blocks are independent matrices.

\subsection{Emergent Expert Path Structure}\label{sec:path_analysis}

We now examine the expert paths that emerge in trained MoE models. If MoE models learn meaningful cross-layer structure, tokens following the same path should share linguistic properties.

\subsubsection{Expert Path Distribution}
\label{sec:path_dist}

For each expert path~$\mathbf{e}$ we define its \textit{frequency} $\text{freq}(\mathbf{e})$ as the proportion of tokens whose top-1 expert selections across layers match $\mathbf{e}$:
\begin{equation}
    \text{freq}(\mathbf{e}) = \frac{|\{x : (E_1(x), \ldots, E_L(x)) = \mathbf{e}\}|}{|\{x\}|}.
\end{equation}

\begin{wrapfigure}{r}{0.4\columnwidth}
    \centering
    \vspace{-1em}
    \includegraphics[width=0.3\columnwidth]{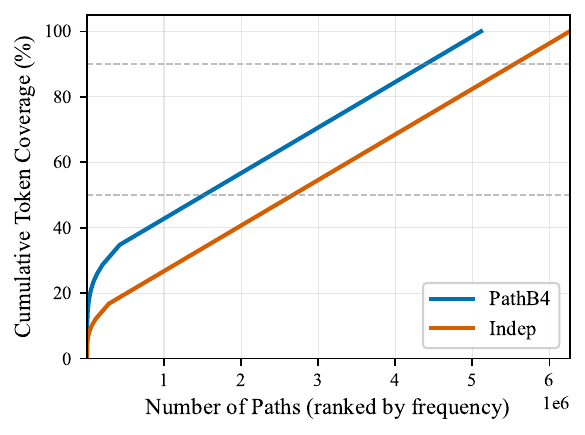}
    \caption{Cumulative token coverage as a function of the number of paths.}
    \label{fig:path_coverage}
    \vspace{-1em}
\end{wrapfigure}

To visualize how tokens are distributed across paths, we sort paths by frequency
and compute the \textit{cumulative token coverage}: the fraction of tokens
covered by the top-$K$ most frequent paths.

\begin{figure*}[t]
    \centering
    \includegraphics[width=\textwidth]{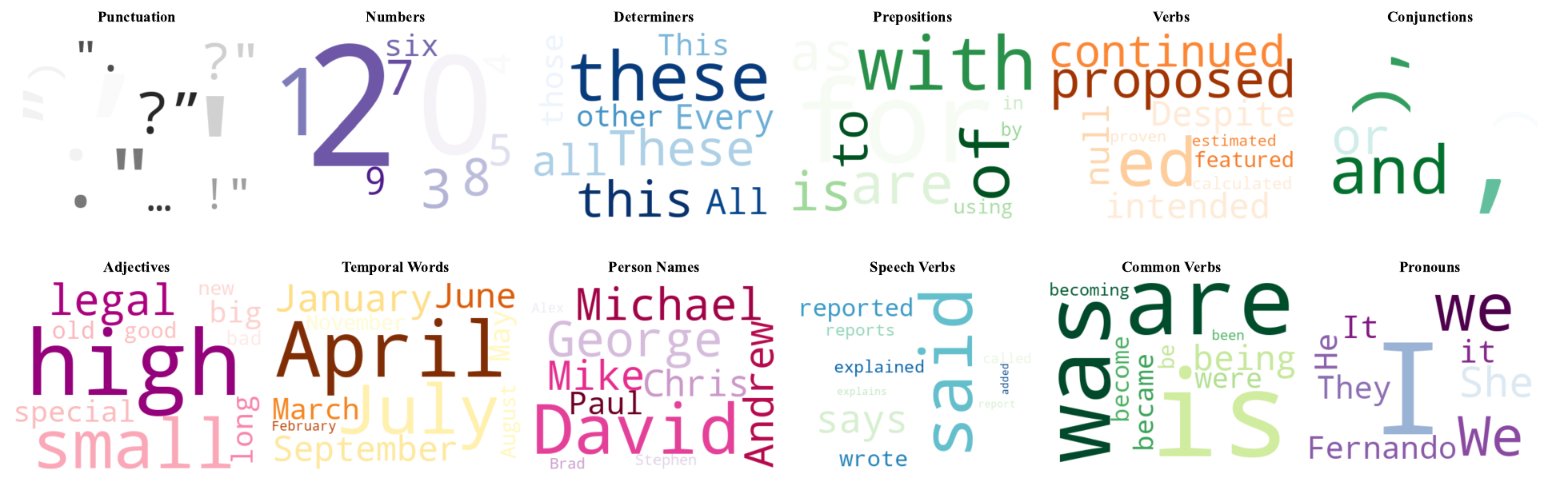}
    \caption{Token specialization of representative expert paths. Each word cloud shows the most frequent tokens processed by a path specialized for that linguistic category.}
    \label{fig:path_specialization}
\end{figure*}

Figure \ref{fig:path_coverage} compares the cumulative token coverage between
\pathmoe{} (with block size 4) and independent routing.
Both models concentrate tokens into a small fraction of the $N^L$ possible paths, confirming that path clustering is an emergent property of MoE routing, not an artifact of our method. \pathmoe{} does so more aggressively, with the top paths covering a larger fraction of tokens.

\subsubsection{Expert Path Token Specialization}
\label{sec:path_spec}

Figure \ref{fig:path_specialization} reveals that expert paths develop distinct linguistic specializations without any supervision.
Some paths predominantly process punctuation, others handle person names or speech verbs (``said'', ``explained''), and yet others specialize in temporal expressions or function words.
This clustering by linguistic function emerges purely from the language modeling objective: the model learns to route syntactically or semantically similar tokens through shared computational pathways.
Notably, this specialization is observed in both independent routing and \pathmoe{}, suggesting it is a general property of MoE routing. However, \pathmoe{} produces more concentrated token clusters within each path.

\subsection{Path Constraints Reduce Entropy}

Because layers in the same block share $\mathbf{W}_b$ and receive similar
representations ($\mathbf{x}_l \approx \mathbf{x}_{l-1}$ via residual
connections), they tend to select the same expert, therefore concentrate tokens into
fewer paths. Independent routing uses separate $\mathbf{W}_l$ per layer,
producing near-independent selections and a much larger effective path space. We
provide an informal analysis via a mutual-information decomposition in
Appendix~\ref{app:entropy_analysis}.

\paragraph{Empirical Routing Entropy.}
We validate this on a network with $L{=}24$ layers and $N{=}16$ experts over ${\sim}$7M tokens. \pathmoe{} ($B{=}4$) achieves 21.14 bits vs.\ 22.20 bits for independent routing, a 1-bit reduction that halves the effective path space. This concentration means each path receives more training signal, improving sample efficiency. Correspondingly, consecutive-layer routing correlation is 85.6\% for \pathmoe{} vs.\ 62\% for independent routing (Appendix~\ref{app:theory_validation}), enabling experts to specialize for coordinated inputs rather than arbitrary representations, yielding faster learning and greater robustness (Section~\ref{sec:analysis}).

\section{Experiments}
\subsection{Experimental Setup}
We evaluate path-constrained routing using the Transformer architecture. Model hyperparameter settings closely follow~\citet{qiu2024layerwise}.

\vspace{-0.3cm}
\paragraph{Training Details.}
Our main experiments use a 0.9B parameter model (0.37B active) with 16 experts and top-4 routing, trained on Fineweb-100B~\citep{penedo2024fineweb} for 400k steps using the Llama 2 tokenizer.
We remove the load balancing loss for \pathmoe{} variants but keep $\alpha=0.01$ for all other models (see discussion in Section \ref{sec:load-balancing}).
Full hyperparameters are given in Appendix~\ref{app:training}.

\begin{table}[t]
  \centering
  \caption{Main results on Fineweb-100B with 0.9B total / 0.37B active MoE architecture. Throughput is reported per GPU and memory reports peak active GPU memory.}
  \label{tab:main_results}
  \resizebox{\textwidth}{!}{
  \begin{tabular}{lcccccccccccc}
      \toprule
      \multirow{2}{*}{\textbf{Routing}} & \multirow{2}{*}{\textbf{ARC-E}} & \multirow{2}{*}{\textbf{BoolQ}} & \multirow{2}{*}{\textbf{HSwag}} & \multirow{2}{*}{\textbf{LAMBADA}} & \multirow{2}{*}{\textbf{OBQA}} & \multirow{2}{*}{\textbf{PIQA}} & \multirow{2}{*}{\textbf{SocIQA}} & \multirow{2}{*}{\textbf{WinoGr.}} & \multirow{2}{*}{\textbf{Avg.}} & \multirow{2}{*}{\textbf{Perplexity}} & \textbf{Throughput} & \textbf{Memory} \\
      & & & & & & & & & & & (k tok/s/GPU) & (GiB) \\
      \midrule
      \rowcolor{gray!10}
      \multicolumn{13}{l}{\textit{Independent Routing}} \\
      Indep-MoE & 44.57 & 56.45 & 45.99 & 46.19 & 29.80 & 66.87 & 38.84 & 51.54 & 47.53 & 12.91 & 55.21 & 66.94 \\
      Rand-MoE & 40.49 & \textbf{61.10} & 37.95 & 38.83 & 28.20 & 64.47 & 38.89 & 50.99 & 45.12 & 16.14 & 75.17 & 47.49 \\
      X-MoE & 43.27 & 60.55 & 46.26 & 45.22 & 32.20 & 67.63 & 39.61 & 52.80 & 48.44 & 13.21 & 56.37 & 67.06 \\
      \midrule
      \rowcolor{blue!5}
      \multicolumn{13}{l}{\textit{Recurrent Routing}} \\
      Recurrent-MoE & 44.07 & 54.50 & 46.05 & 47.12 & 30.80 & \textbf{67.74} & 38.79 & 53.67 & 47.84 & 12.92 & 54.65 & 69.26 \\
      \midrule
      \rowcolor{green!5}
      \multicolumn{13}{l}{\textit{Path-Constrained Routing}} \\
      MonoB8-MoE & 42.55 & 57.80 & 46.36 & 46.73 & 31.20 & 65.67 & 37.56 & 54.06 & 47.74 & 13.34 & 56.53 & 62.86 \\
      LowRank-MoE & 44.36 & 59.97 & 46.33 & 46.11 & 31.80 & 66.65 & 39.20 & 52.96 & 48.42 & 12.98 & 53.84 & 66.96 \\
      \addlinespace[3pt]
      PathMoE & 45.50 & 55.44 & 47.82 & 47.89 & \textbf{32.40} & 66.21 & 38.49 & 53.43 & 48.40 & 12.92 & 55.71 & 66.74 \\
      PathB8-MoE & 43.73 & 59.97 & 46.65 & 46.92 & 31.40 & 66.81 & 40.69 & 54.93 & 48.89 & 12.53 & 55.20 & 66.94 \\
      PathB4-MoE & \textbf{44.70} & 60.40 & \textbf{47.95} & \textbf{49.45} & 31.60 & 66.32 & \textbf{40.94} & \textbf{55.64} & \textbf{49.62} & \textbf{12.29} & 55.67 & 66.83 \\
      \bottomrule
  \end{tabular}
  \vspace{-0.3cm}
}
\end{table}

\subsection{Performance Comparison}
\label{sec:comparison}

\paragraph{Baseline Methods.}
We compare against three categories.
\textit{Independent routing}: \textbf{Indep-MoE} (standard per-layer routers), \textbf{Rand-MoE} (frozen random routers), and \textbf{X-MoE}~\citep{chi2022representation} (routing in a normalized low-dimensional space).
\textit{Recurrent routing}: \textbf{Recurrent-MoE}~\citep{qiu2024layerwise} (GRU-based router conditioned on previous layers).
\textit{Path-constrained routing}: \textbf{LowRank-MoE} (shared base router + low-rank per-layer perturbation), \textbf{Mono-MoE} (shared routing \emph{decisions} within blocks), and \textbf{\pathmoe{}} (shared routing \emph{parameters} within blocks). See Appendix~\ref{app:baselines} for detailed formulations.
For block-based methods, B$n$ denotes block size $n$; omitting the suffix means sharing across all layers.

\vspace{-0.3cm}
\paragraph{Evaluation Tasks.}
We evaluate on eight downstream tasks: ARC-Easy~\citep{clark2018think}, BoolQ~\citep{clark2019boolq}, HellaSwag~\citep{zellers2019hellaswag}, LAMBADA~\citep{paperno2016lambada}, OpenBookQA~\citep{mihaylov2018can}, PIQA~\citep{bisk2020piqa}, SocialIQA~\citep{sap2019socialiqa}, and WinoGrande~\citep{sakaguchi2020winogrande}.
All results report accuracy (\%); OpenBookQA, PIQA, ARC-Easy, and HellaSwag use length-normalized accuracy.

\vspace{-0.3cm}
\paragraph{Results.}
Table \ref{tab:main_results} shows results for all models trained under identical compute budgets (400k steps on Fineweb-100B, best of final 5 checkpoints).

\textbf{Perplexity.} PathB4-MoE, the best-performing configuration for our architecture (see Section~\ref{sec:ablation} for block size ablations), achieves the lowest perplexity (12.29 vs.\ 12.91
for Indep-MoE) over $\sim$ 7M test tokens,
indicating that path constraints lead to better language modeling.

\textbf{Downstream tasks.} PathB4-MoE achieves the highest average accuracy
(49.62\%), with gains across most benchmarks. Sharing parameters (\pathmoe{})
consistently outperforms sharing decisions (Mono-MoE), confirming that soft constraints via shared parameters are more effective than hard constraints that force identical selections.

\textbf{Generalization.} \pathmoe{} provides orthogonal benefits on top of X-MoE
(Appendix~\ref{app:orthogonal}). These advantages also persist beyond the
100B-token regime: on DCLM-Pro with 5$\times$ more data (500B tokens),
\pathmoe{} still outperforms Indep-MoE (Appendix~\ref{app:dclm1b}), suggesting
that the gains stem from a better inductive bias rather than data efficiency
alone.

\vspace{-0.3cm}
\paragraph{Efficiency.} PathB4-MoE matches Indep-MoE in throughput (55.67 vs.\ 55.21 k tok/s/GPU) and memory (66.83 vs.\ 66.94 GiB), since the only change is sharing router parameters, which reduces parameter count while keeping identical
forward and backward computation.

\subsection{Load Balancing Without Auxiliary Losses}\label{sec:load-balancing}

\begin{wrapfigure}{r}{0.5\columnwidth}
  \centering
  \vspace{-1.5em}
  \includegraphics[width=0.5\columnwidth]{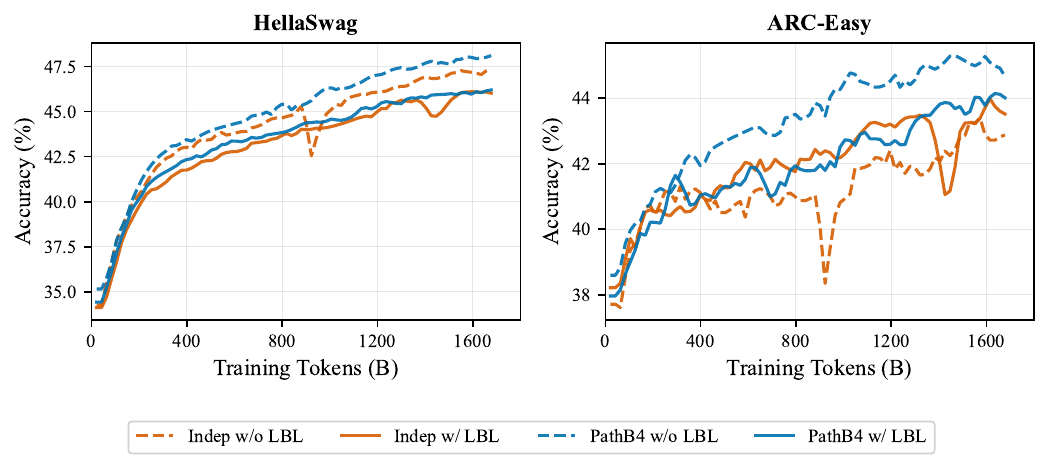}
  \caption{Training dynamics with and without load balancing losses (LBL).}
  \label{fig:learning_curves}
  \vspace{-1em}
\end{wrapfigure}

Conventional MoE training employs auxiliary load balancing losses to prevent expert collapse~\citep{fedus2022switch}.
We find that \pathmoe{} can be trained without auxiliary losses.
Figure \ref{fig:learning_curves} shows training dynamics of PathB4-MoE and Indep-MoE on HellaSwag and ARC-Easy, with and without load balancing losses (LBL).
PathB4-MoE without LBL (dashed blue) achieves the best final accuracy on both tasks, while removing LBL from Indep-MoE leads to more erratic training dynamics.
This suggests that \pathmoe{} is more robust to the removal of auxiliary losses,
maintaining smooth convergence regardless of the load balancing setting.
Additionally, the accuracy gap between \pathmoe{} and Indep-MoE remains stable
throughout training, suggesting the benefit is architectural rather than a
transient data-efficiency effect that diminishes as routers receive more
training signal. See full results in Appendix~\ref{app:lbl}.

\subsection{Scaling to 16B Parameters}
\label{sec:scaling}

\begin{wraptable}{r}{0.48\columnwidth}
  \centering
  \vspace{-3em}
  \caption{16B model results (16.2B total / 2.13B active) on DCLM-Pro.}
  \label{tab:16b_results}
  \scriptsize
  \begin{tabular}{llccr}
      \toprule
      & \textbf{Task} & \textbf{Indep} & \textbf{Path} & \textbf{$\Delta$} \\
      \midrule
      \multirow{4}{*}{\textit{Comm.}}
      & WinoGr. & 56.12 & \textbf{57.93} & \textcolor{green!50!black}{+1.81} \\
      & PIQA & 68.99 & \textbf{71.38} & \textcolor{green!50!black}{+2.39} \\
      & SocIQA & \textbf{40.58} & 40.07 & \textcolor{red!70!black}{--0.51} \\
      & CSQA & 43.00 & \textbf{48.73} & \textcolor{green!50!black}{+5.73} \\
      \midrule
      \multirow{3}{*}{\textit{Lang.}}
      & LAMBADA & 51.97 & \textbf{53.64} & \textcolor{green!50!black}{+1.67} \\
      & HSwag & 63.22 & \textbf{63.32} & \textcolor{green!50!black}{+0.10} \\
      & BoolQ & 63.06 & \textbf{64.01} & \textcolor{green!50!black}{+0.95} \\
      \midrule
      \multirow{5}{*}{\textit{Knowl.}}
      & OBQA & 35.80 & \textbf{39.60} & \textcolor{green!50!black}{+3.80} \\
      & ARC-E & 63.43 & \textbf{68.52} & \textcolor{green!50!black}{+5.09} \\
      & ARC-C & 37.29 & \textbf{40.10} & \textcolor{green!50!black}{+2.81} \\
      & TriviaQA & \textbf{21.91} & 21.18 & \textcolor{red!70!black}{--0.73} \\
      & MMLU & 35.27 & \textbf{35.64} & \textcolor{green!50!black}{+0.37} \\
      \midrule
      & \textbf{Avg.} & 48.39 & \textbf{50.34} & \textcolor{green!50!black}{+1.95} \\
      \bottomrule
  \end{tabular}
  \vspace{-1em}
\end{wraptable}

To verify that path-constraining benefits persist at larger scales, we train 16B parameter models (16.2B total, 2.13B active) with 64 experts and top-6 routing on DCLM-Pro~\citep{zhou2024programming} for 200k steps using the GPT-NeoX-20B tokenizer, without load balancing loss.
We additionally evaluate on MMLU~\citep{hendrycks2020measuring}, ARC-Challenge~\citep{clark2018think}, CommonsenseQA~\citep{talmor2019commonsenseqa}, and TriviaQA~\citep{joshi2017triviaqa} (5-shot exact match). Full hyperparameters are in Appendix~\ref{app:training}. Table \ref{tab:16b_results} shows that PathMoE wins on 10 of 12 tasks with +1.95\% average improvement.

\subsection{Expert Path Concentration}
\label{sec:path_conc}

To understand the relationship between path concentration and model performance, we evaluate models with different levels of path restriction.
Notably, the dominant paths emerge very early in training: we train \pathmoe{} for only 5k steps (1.25\% of total training), save a checkpoint, and identify the most frequent paths at this early stage. We then continue training with routing restricted to these top-10, top-100, or top-500 paths.
Restricting to only 10 paths significantly degrades performance (46.22\% avg.), 100 paths recovers most of the performance (48.18\%), and 500 paths (48.86\%) performs comparably to unrestricted routing (48.71\%). This suggests the model naturally concentrates on a moderate number of specialized paths. Full per-task results are in Appendix~\ref{app:path_concentration}.

\section{Ablation Studies}
\label{sec:ablation}

\begin{figure}[h!]
    \centering
    \includegraphics[width=0.7\columnwidth]{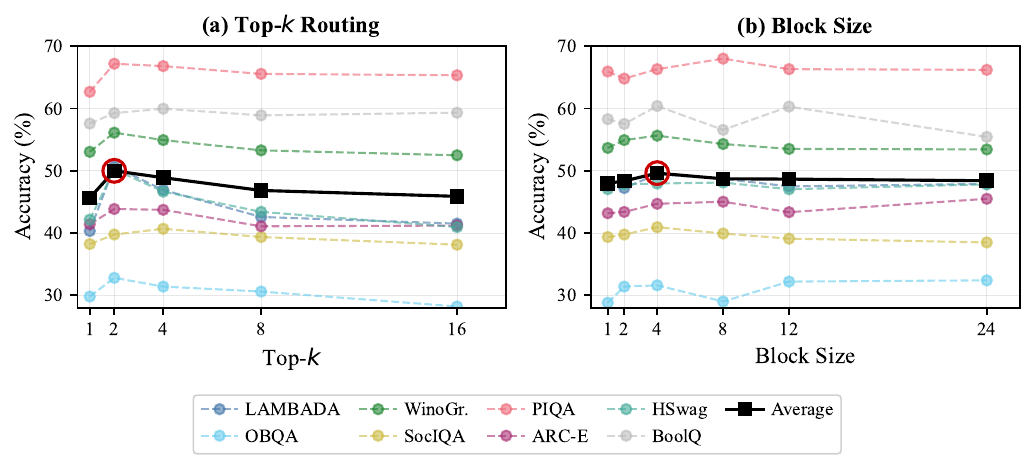}
    \caption{Ablation studies on routing hyperparameters. (a) Effect of top-$k$ routing. (b) Effect of block size. Dashed lines show individual benchmarks; solid black line shows the average.}
    \label{fig:ablation}
\end{figure}

\vspace{-0.3cm}
\paragraph{Top-$k$ Routing Analysis.}
We investigate the effect of the top-$k$ routing parameter on the 0.9B architecture across values of 1, 2, 4, 8, and 16.
Figure \ref{fig:ablation}(a) shows that top-2 routing achieves the best average accuracy (49.99\%), consistent with Mixtral~\citep{jiang2024mixtral}.
Performance degrades at both extremes: $k$=1 suffers from training instability and limited capacity, while $k$=16 dilutes expert contributions---both yield similar accuracy ($\sim$45.7\%).
We attribute the advantage of $k$=2 to the balance between two competing effects.
Increasing $k$ allows the model to combine complementary expert computations, improving representational capacity.
However, larger $k$ weakens the routing signal per expert: when many experts contribute, each receives a smaller share of the gradient, reducing specialization pressure.
At $k$=2, the model can leverage one primary expert for the dominant computation and a secondary expert for residual corrections, without diluting the routing learning signal.

\vspace{-0.3cm}
\paragraph{Block Size Analysis.}
We investigate the effect of block size on the 0.9B architecture across values of 1, 2, 4, 8, 12, and 24 layers.
Block size controls the trade-off between routing flexibility and cross-layer coordination: small blocks (B1, B2) provide insufficient coordination as layers learn nearly independently, while large blocks (B12, B24) over-constrain the router to handle representations that change substantially across layers.
Figure \ref{fig:ablation}(b) shows that intermediate sizes ($B=4$) work best for the
architecture we experiment.
We also observe task-dependent preferences---language modeling and commonsense tasks
peak at B4, physical reasoning at B8, and knowledge-intensive tasks at
B24, suggesting that \textit{the optimal block size may vary by domain}.
More broadly, the optimal block size will likely vary across model architectures and depths. Nonetheless, path-constrained routing consistently improves over independent routing across all block sizes we tested.

\section{Understanding Path Constraints: Routing Consistency and Robustness}
\label{sec:analysis}

Beyond performance gains, we examine cross-layer routing
consistency and robustness to perturbations to understand the effects of path constraints can guide future MoE design.

\vspace{-0.3cm}
\subsection{Cross-Layer Routing Consistency}

Since expert indices are arbitrary across layers (expert 0 at layer $l$ has no inherent correspondence to expert 0 at layer $l+1$), we first align indices using Algorithm~\ref{alg:alignment}.

\begin{wrapfigure}{r}{0.45\columnwidth}
\vspace{-1.5em}
\begin{minipage}{0.45\columnwidth}
\begin{algorithm}[H]
\caption{Expert Index Alignment}\label{alg:alignment}
\small
\begin{algorithmic}[1]
\Require Top-$k$ expert assignments for all tokens across $L$ layers
\For{each consecutive layer pair $(l, l+1)$}
    \State Construct co-occurrence matrix $C \in \mathbb{R}^{N \times N}$, where $C_{ij} = \sum_t \sum_{a \in \text{top-}k(t,l)} \sum_{b \in \text{top-}k(t,l+1)} \mathbf{1}[a{=}i, b{=}j]$
    \State Find $\sigma^*$ = \Call{Hungarian}{$C$} that maximizes $\sum_i C_{i,\sigma(i)}$
    \State Relabel layer $l+1$: $i \leftarrow (\sigma^*)^{-1}(i)$
\EndFor
\end{algorithmic}
\end{algorithm}
\end{minipage}
\vspace{-1em}
\end{wrapfigure}
We then examine two complementary metrics.
\textbf{Path consistency} measures expert reuse: for a given window size $w$, we compute the average Jaccard similarity between expert sets at consecutive layers within sliding windows of $w$ layers.
\textbf{Sustained engagement} measures how long a token continues using a given
aligned expert: we count runs of consecutive layers and report the fraction
lasting $\geq$X layers.

\begin{figure*}[h!]
  \centering
  \includegraphics[width=0.48\textwidth]{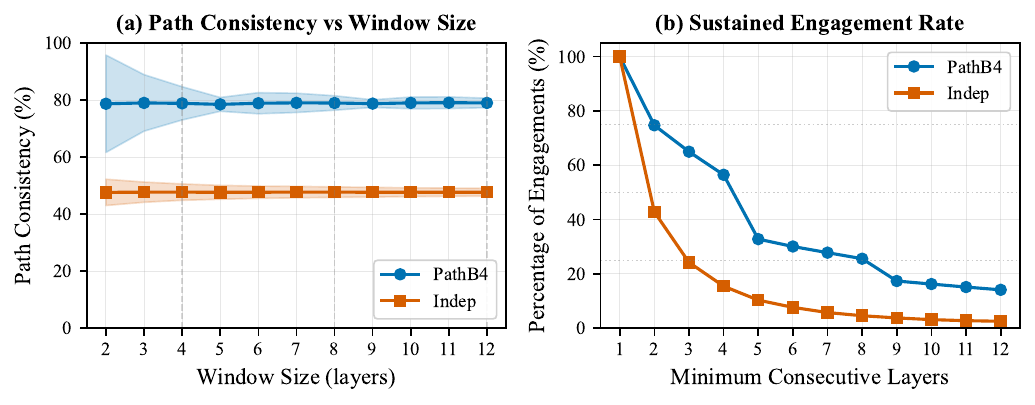}
  \hfill
  \includegraphics[width=0.48\textwidth]{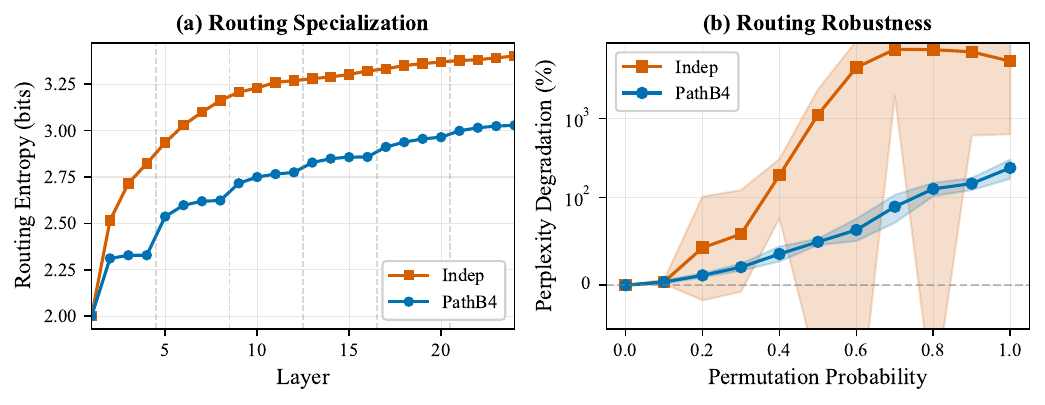}
  \caption{Routing consistency, specialization, and robustness. (a) Path consistency versus window size. (b) Sustained engagement versus minimum consecutive layers. (c) Cumulative routing entropy across layers. (d) Perplexity degradation under expert permutation.}
  \label{fig:consistency_robustness}
\end{figure*}

Figure \ref{fig:consistency_robustness}(a) shows that PathB4-MoE maintains consistently higher path consistency ($\sim$79\%) compared to Indep-MoE ($\sim$48\%) across all window sizes from 2 to 12 layers.
This gap persists even at non-block-aligned windows (3, 5, 7 layers), confirming the benefit stems from the architecture rather than block boundaries.
Panel (b) measures how long tokens continue using the same expert, where
\pathmoe{} produces substantially more sustained engagements.

\vspace{-0.3cm}
\subsection{Expert Specialization and Routing Robustness}

We examine the specialization-robustness trade-off to check whether \pathmoe{}'s
routing creates brittle dependencies.

\vspace{-0.3cm}
\paragraph{Routing Specialization.}
We measure specialization via cumulative routing entropy: at each layer $l$, we compute the entropy of each token's expert usage distribution from layer 1 to $l$, where lower entropy indicates more concentrated (specialized) routing.
Figure \ref{fig:consistency_robustness}(c) shows that PathB4-MoE achieves consistently lower entropy than Indep-MoE across all layers, with 11\% lower entropy at the final layer.

\vspace{-0.3cm}
\paragraph{Routing Robustness.}
We test robustness by randomly permuting expert indices at each layer with probability $p$ and measuring perplexity degradation.
Figure \ref{fig:consistency_robustness}(d) shows that PathB4-MoE is dramatically
more robust despite being more specialized: at full permutation, Indep-MoE
degrades by 5,328\% versus only 237\% for PathB4-MoE. This suggests that path-constrained routing learns more distributed representations within each expert, rather than relying on fragile per-layer specializations.

\section{Related Work}

\vspace{-0.3cm}
\paragraph{MoE Architectures.}
The concept of MoE dates back to~\citet{jacobs1991adaptive}, but has seen renewed interest with the scaling of deep learning models.~\citet{shazeer2017outrageously} introduced sparsely-gated MoE layers for neural machine translation, demonstrating that conditional computation could dramatically increase model capacity with modest computational overhead.
This work has been extended to large language models with architectures such as Switch Transformers~\citep{fedus2022switch}, GLaM~\citep{du2022glam}, and recently, Mixtral~\citep{jiang2024mixtral} and DeepSeek-MoE~\citep{dai2024deepseekmoe}.
While these architectures have demonstrated impressive scaling properties, they primarily employ independent routing at each layer, treating expert selection as a per-layer decision without considering cross-layer structure.

\vspace{-0.3cm}
\paragraph{Routing Mechanisms in MoE.}
The routing function is critical to MoE performance, determining how inputs are distributed to experts.
Classical approaches employ top-$k$ routing with load balancing constraints~\citep{shazeer2017outrageously, lepikhin2020gshard}.
Recent work has explored various improvements:~\citet{raposo2024mixture} proposed mixture-of-depths that dynamically adjusts computation depth;~\citet{dai2022stablemoe} introduced deterministic routing for training stability;~\citet{zhou2022mixture} proposed expert choice routing where experts select tokens;
and~\citet{puigcerver2023sparse} investigated soft mixtures that combine all experts with learned weights.
These approaches improve individual routing decisions but do not address
cross-layer coordination, which our analysis suggests is important for path efficiency.

\vspace{-0.3cm}
\paragraph{Expert Specialization and Redundancy.}
A growing body of work has investigated what experts learn and whether they develop meaningful specializations.~\citet{chi2022representation} showed that sparse MoE can suffer from representation collapse where experts learn redundant functions.~\citet{yang2024moe} proposed compression techniques exploiting inter-expert redundancy.~\citet{zadouri2023pushing} proposed methods to encourage expert diversity through regularization.
Recently,~\citet{antoine2024pos} analyzed routing paths across six MoE architectures and found that experts specialize for specific part-of-speech categories, with routing paths showing high predictive accuracy for POS tags---providing independent evidence that routing paths encode meaningful linguistic structure.
Path-constrained routing addresses these concerns architecturally: by encouraging tokens
to follow consistent expert paths, experts naturally specialize for different
token types without explicit regularization.

\vspace{-0.3cm}
\paragraph{Cross-Layer Routing Coordination.}
While most MoE architectures employ independent routers per layer, recent work has begun exploring cross-layer coordination.~\citet{qiu2024layerwise} proposed a recurrent router that conditions each layer's routing on previous decisions.~\citet{gu2025omni} demonstrated that sharing router parameters across all layers improves speech recognition, where acoustic features follow structured path patterns.
Orthogonally,~\citet{dai2022stablemoe} identified routing fluctuation across training steps and proposed distilling a stable router, addressing temporal rather than spatial coordination.
Our work builds on these insights by proposing the expert path perspective as an analytical framework, with block-wise parameter sharing (\pathmoe{}) as one instantiation that balances path consistency with representational flexibility across network depth.

\section{Conclusion}
\vspace{-0.3cm}

We proposed viewing MoE architectures through the lens of \emph{expert paths}---the cross-layer sequences of expert selections that tokens follow.
This perspective reveals that tokens naturally cluster into a small, interpretable subset of paths aligned with linguistic function, despite the combinatorial size of the path space.

As one instantiation of path-constrained design, we introduced \pathmoe{}, which shares router parameters within blocks of consecutive layers to constrain the path space.
Experiments on 0.9B and 16B parameter models show consistent performance gains (+2.1\% average accuracy on 0.9B; 10/12 tasks improved on 16B) without auxiliary load balancing losses, along with improved cross-layer coordination (79\% vs.\ 48\% routing consistency) and more specialized yet robust routing (11\% lower entropy, 22.5$\times$ more robust to perturbations).

These results establish expert paths as a useful design axis for MoE architectures, complementary to existing work on per-layer routing mechanisms.
\pathmoe{} is one way to constrain the path space; we expect this perspective to motivate further architectural designs beyond parameter sharing.

\section{Limitations}
\vspace{-0.3cm}

Block-wise parameter sharing assumes token-choice routing; preliminary experiments show no benefit for expert-choice routing, which already achieves consistency through stable expert preferences.
Our experiments are conducted at 0.9B and 16B total parameters; validating path constraints at larger scales remains future work due to computational constraints.
Additionally, while we explore block-wise parameter sharing as one instantiation, other mechanisms for constraining the path space (e.g., learned path predictors, path-aware regularization) remain unexplored.

\section*{Impact Statement}
This paper introduces an architectural modification to Mixture-of-Experts routing that improves downstream model performance, cross-layer coordination and model interpretability. By encouraging tokens to follow coherent expert paths, our approach reveals how MoE models naturally develop linguistic specializations---providing insights that could aid model understanding and debugging.

The efficiency gains from eliminating auxiliary load balancing losses and achieving better parameter utilization could reduce the computational cost of training large models. However, our work focuses on routing mechanisms rather than capability scaling, and does not introduce risks beyond those inherent to large language models research.

\section*{Acknowledgements}
We thank Ronan Collobert, Yizhe Zhang, Samira Abnar, Anastasiia Filippova,
Shuangfei Zhai, Russ Webb and Barry Theobald for helpful discussions and
feedback.

\bibliographystyle{plainnat}
\bibliography{reference}

\newpage
\appendix

\section{Routing Entropy Analysis}
\label{app:entropy_analysis}

We provide an informal approximation of routing entropy to illustrate why
\pathmoe{} routing may reduce the effective size of the expert path space
compared to independent routing.\footnote{Note that: (i) in the worst case,
\pathmoe{} can degenerate to uniform expert routing (e.g. $\mathbf{W}_b=0$);
(ii) in the best case, independent routing can learn the same $\mathbf{W}_l$
within the same block.}

Using the chain rule of entropy, the routing entropy can be decomposed and further approximated using the first-order Markov dependency (meaningful approximation in residual networks):
\begin{equation*}
    H(\mathbf{E}) = \sum_{l=1}^L H(E_l \mid E_1 \ldots E_{l-1}) \approx \sum_{l=1}^L H(E_l \mid E_{l-1}).
\end{equation*}
Let $\mathcal{S}$ be the set of layer indices that start a new block, $H_{\text{indep}}(\mathbf{E})$ be the independent routing entropy and $H_{\pathmoe{}}(\mathbf{E})$ be the \pathmoe{} routing entropy. Then
\begin{align*}
\Delta H = H_{\text{indep}}(\mathbf{E}) - H_{\pathmoe{}}(\mathbf{E}) \\
\approx \sum_{l\notin \mathcal{S}} [H_{\text{indep}}(E_l \mid E_{l-1}) - H_{\pathmoe{}}(E_l \mid E_{l-1})].
\end{align*}
Using the definition of conditional entropy $H(E_l \mid E_{l-1}) = H(E_l) - I(E_l; E_{l-1})$,
where $I(E_l; E_{l-1})$ is the mutual information between consecutive layers, we have:\footnote{Every entropy is defined on the marginal distribution of the corresponding random variable over the data distribution $\mathcal{D}$.}
\begin{align*}
\Delta H \approx \sum_{l\notin \mathcal{S}} [I_{\pathmoe{}}(E_l; E_{l-1}) - I_{\text{indep}}(E_l; E_{l-1})].
\end{align*}
Given the specific representations $\mathbf{x}_l$ and $\mathbf{x}_{l-1}$, the routing decisions are sampled independently via the stochastic mechanism:
\begin{equation*}
    P(E_l, E_{l-1} \mid \mathbf{x}_l, \mathbf{x}_{l-1}) = P(E_l \mid \mathbf{x}_l) P(E_{l-1} \mid \mathbf{x}_{l-1}).
\end{equation*}
Let $p^{(l)}_k(\mathbf{x}_l) = P(E_l=k \mid \mathbf{x}_l)$ be the probability function for expert $k$.
The marginal joint distribution is defined as:
\begin{equation*}
    P(E_l=i, E_{l-1}=j) = \mathbb{E}_{\mathbf{x} \sim \mathcal{D}} \left[ p^{(l)}_i(\mathbf{x}_l) \cdot p^{(l-1)}_j(\mathbf{x}_{l-1}) \right].
\end{equation*}
Then the layers are independent only if the routing probabilities are uncorrelated across the data.

\textit{Independent Routing.}
Given that (i) router parameters $\mathbf{W}_l$ and $\mathbf{W}_{l-1}$ are initialized independently and stay independent for some time during training; (ii) due to residual connections representations evolve slowly $\mathbf{x}_l\approx \mathbf{x}_{l-1}$; the decision boundaries of routing function at layer $l$ are statistically orthogonal to those at layer $l-1$ across the data. Thus
\begin{align*}
    P(E_l=i, E_{l-1}=j) \approx P(E_l=i) P(E_{l-1}=j)  \\
    = \mathbb{E}_{\mathbf{x} \sim \mathcal{D}} \left[ p^{(l)}_i(\mathbf{x}_l) \right] \mathbb{E}_{\mathbf{x} \sim \mathcal{D}} \left[ p^{(l-1)}_j(\mathbf{x}_{l-1}) \right] \\
    \implies I_{\text{indep}}(E_l; E_{l-1}) \approx 0.
\end{align*}
\textit{\pathmoe{} Routing.}
As layers $l, l-1$ are within a shared block, the router parameters are identical $\mathbf{W}_l = \mathbf{W}_{l-1}$. Due to residual connections, representations evolve slowly $\mathbf{x}_l\approx \mathbf{x}_{l-1}$, thus routing probability functions are close:
$p^{(l)}_k(\mathbf{x}_l) \approx p^{(l-1)}_k(\mathbf{x}_{l-1}) \triangleq \phi_k(\mathbf{x}_l).$
Then the joint probability of selecting the same expert $k$ at both layers:
\begin{equation*}
    P(E_l=k, E_{l-1}=k) \approx \mathbb{E}_{\mathbf{x}\sim\mathcal{D}} \left[ \phi_k(\mathbf{x}_l)^2 \right].
\end{equation*}
By Jensen's inequality, since $f(z)=z^2$ is strictly convex and the routing probability $\phi_k(\mathbf{x}_l)$ varies across the data distribution, we get that $E_l$ and $E_{l-1}$ are correlated and this correlation is concentrated on the diagonal:
\begin{align*}
    \mathbb{E}_{\mathbf{x}\sim\mathcal{D}} \left[ \phi_k(\mathbf{x}_l)^2 \right] > \left( \mathbb{E}_{\mathbf{x}\sim\mathcal{D}} [\phi_k(\mathbf{x}_l)] \right)^2 \\
    P(E_l=k, E_{l-1}=k)> P(E_l=k)\cdot P(E_{l-1}=k) \\
    \implies P(E_l, E_{l-1})\neq P(E_l)\cdot P(E_{l-1}) \\
    \implies I_{\pathmoe{}}(E_l; E_{l-1}) \gg 0.
\end{align*}
Given $I_{\pathmoe{}}(E_l;E_{l-1}) \gg I_{\text{indep}}(E_l; E_{l-1}) \approx 0$, we obtain ${\Delta H > 0}$. This shows that \pathmoe{} may introduce strong inter-layer correlations for the marginal distribution of expert paths over the data distribution $\mathcal{D}$, significantly constraining the effective expert path space.

\section{Empirical Routing Entropy}
\label{app:theory_validation}

We provide detailed empirical routing entropy computation and expert path space
statistics for both independent routing and \pathmoe{} routing from
Section~\ref{sec:theory}. Table \ref{tab:theory_validation_appendix} reports
comprehensive metrics for Indep-MoE and PathB4-MoE measured on 7.18M tokens.

\paragraph{Estimation Procedure.}
For each token in the evaluation set, we record the top-1 expert selection at
each of the $L=24$ layers, producing an expert path $\mathbf{e} = (e_1, \ldots, e_L)
\in [N]^L$.
We count the empirical frequency of each observed path using a hash map over the 7.18M tokens, yielding the empirical distribution $\hat{P}(\mathbf{e}) = \text{count}(\mathbf{e}) / T$ where $T$ is the total number of tokens.
Despite the theoretical support of $16^{24} \approx 10^{29}$ paths, the vast majority are never observed; only ${\sim}$5--6M unique paths appear, making direct enumeration over observed paths tractable.
The empirical routing entropy is then computed as $\hat{H}(\mathbf{E}) = -\sum_{\mathbf{e}: \text{count}(\mathbf{e})>0} \hat{P}(\mathbf{e}) \log_2 \hat{P}(\mathbf{e})$, summing only over observed paths.
The consecutive-layer correlation is computed as the fraction of tokens for which $e_l = e_{l+1}$ (after expert index alignment), averaged over all consecutive layer pairs.

\begin{table}[h]
  \centering
  \caption{Detailed empirical routing entropy computation and expert path space statistics for Indep-MoE and PathB4-MoE (block size $B=4$) with $L=24$ layers and $N=16$ experts. All entropy measurements are in bits. We use evaluation data with 7.18M tokens.}
  \label{tab:theory_validation_appendix}
  \begin{tabular}{lcc}
      \toprule
      \textbf{Empirical Metric} & \textbf{PathB4-MoE} & \textbf{Indep-MoE} \\
      \midrule
      Routing entropy $H(\mathbf{E})$ & 21.14 bits & 22.20 bits \\
      Routing decisions correlation for consecutive layers & 85.6\% & 62\% \\
      Unique paths observed & 5,109,282 & 6,263,708\\
      Effective path space $2^{H(\pi)}$ & $2.31 \times 10^6$ & $4.82 \times
      10^6$\\
      \bottomrule
  \end{tabular}
\end{table}

\section{Training Details}
\label{app:training}

Table \ref{tab:hyperparameters} summarizes the hyperparameters for both model scales.

\begin{table}[h!]
  \centering
  \caption{Training hyperparameters for 0.9B and 16B models.}
  \label{tab:hyperparameters}
  \begin{tabular}{lcc}
      \toprule
      \textbf{Hyperparameter} & \textbf{0.9B Model} & \textbf{16B Model} \\
      \midrule
      \multicolumn{3}{l}{\textit{Model Architecture}} \\
      Total parameters & 0.9B & 16.2B \\
      Active parameters & 0.37B & 2.13B \\
      Model dimension & 1280 & 2048 \\
      FFN hidden dimension & 768 & 2112 \\
      Layers & 24 & 28 \\
      Attention heads & 20 & 16 \\
      KV heads & 4 & 16 \\
      Experts & 16 & 64 \\
      Top-$k$ routing & 4 & 6 \\
      \midrule
      \multicolumn{3}{l}{\textit{Training}} \\
      Dataset & Fineweb-100B & DCLM-Pro \\
      Tokenizer & Llama 2 (32k) & GPT-NeoX-20B \\
      Sequence length & 4096 & 4096 \\
      Global batch size & 1024 & 2048 \\
      Training steps & 400k & 200k \\
      \midrule
      \multicolumn{3}{l}{\textit{Optimization}} \\
      Optimizer & AdamW & AdamW \\
      $\beta_1, \beta_2$ & 0.9, 0.95 & 0.9, 0.95 \\
      Weight decay & 0.1 & 0.1 \\
      Peak learning rate & $4.2 \times 10^{-4}$ & $2.0 \times 10^{-4}$ \\
      LR schedule & Cosine & Cosine \\
      Warmup steps & 2000 & 2000 \\
      Precision & BF16 & BF16 \\
      \bottomrule
  \end{tabular}
\end{table}

The 0.9B model architecture follows~\citet{qiu2024layerwise}. The 16B model architecture follows DeepSeek-MoE~\citep{dai2024deepseekmoe} without shared experts to isolate the effect of routing changes.

\section{Baseline Descriptions}
\label{app:baselines}

We provide detailed descriptions of the path-constrained routing baselines used
in Section~\ref{sec:comparison}.

\paragraph{LowRank-MoE.} All layers share a base router matrix $\mathbf{W}_{\text{base}}$, and each layer $l$ adds a low-rank perturbation:
\begin{equation}
    \mathbf{W}_l = \mathbf{W}_{\text{base}} + \mathbf{A}_l \mathbf{B}_l,
\end{equation}
where $\mathbf{W}_{\text{base}} \in \mathbb{R}^{N \times d}$ is shared across all layers, and $\mathbf{A}_l \in \mathbb{R}^{N \times r}$, $\mathbf{B}_l \in \mathbb{R}^{r \times d}$ are per-layer low-rank factors with rank $r \ll d$. This allows each layer to deviate slightly from the shared routing while keeping most parameters tied. We use $r = 16$ in our experiments.

\paragraph{Mono-MoE.} Within each block of $B$ consecutive layers, a single routing \emph{decision} is made at the first layer and reused for all subsequent layers in the block:
\begin{equation}
    E_l = E_{\lfloor (l-1)/B \rfloor \cdot B + 1}, \quad \forall l \text{ in the same block}.
\end{equation}
Unlike \pathmoe{}, which shares router \emph{parameters} but allows different decisions based on evolving representations, Mono-MoE forces identical expert selections within each block.

\section{Additional Results}
\subsection{Results on removing load balancing loss (0.9B Model)}
\label{app:lbl}
We provide full experimental results on keeping and removing load balancing loss (LBL) for Indep-MoE and PathB4-MoE.

\begin{table}[h!]
  \centering
  \caption{Performance comparison on Fineweb-100B dataset (0.9B total / 0.37B active).}
  \label{tab:lbl_comparison}
  \resizebox{\columnwidth}{!}{%
  \begin{tabular}{lcccccccccc}
      \toprule
      \textbf{Routing} & \textbf{LBL} & \textbf{ARC-E} & \textbf{BoolQ} & \textbf{HSwag} & \textbf{LAMBADA} & \textbf{OBQA} & \textbf{PIQA} & \textbf{SocIQA} & \textbf{WinoGr.} & \textbf{Avg.}\\
      \midrule
      Indep-MoE & 0.01 & 44.57 & 56.45 & 45.99 & 46.19 & 29.80 & 66.87 & 38.84 & 51.54 & 47.53 \\
      Indep-MoE & 0 & 43.18 & 58.29 & 47.05 & 47.55 & 28.80 & 65.94 & 39.36 & 53.67 & 47.98 \\
      PathB4-MoE & 0.01 & 44.15 & 57.06 & 46.07 & 46.61 & 32.60 & 66.49 & 40.23 & 51.93 & \textbf{48.14} \\
      PathB4-MoE & 0 & 44.70 & 60.40 & 47.95 & 49.45 & 31.60 & 66.32 & 40.94 & 55.64 & \textbf{49.62} \\
      \bottomrule
  \end{tabular}
  }
\end{table}

\subsection{Results on DCLM-Pro Dataset (0.9B Model)}
\label{app:dclm1b}

We provide additional experimental results on the DCLM-Pro dataset using the 0.9B (total) parameter model. Table \ref{tab:dclm_1b} presents the performance comparison across routing methods.

\begin{table}[h!]
  \centering
  \caption{Performance comparison on DCLM-Pro dataset (0.9B total / 0.37B active).}
  \label{tab:dclm_1b}
  \resizebox{\columnwidth}{!}{%
  \begin{tabular}{lccccccccc}
      \toprule
      \textbf{Routing} & \textbf{ARC-E} & \textbf{BoolQ} & \textbf{HSwag} & \textbf{LAMBADA} & \textbf{OBQA} & \textbf{PIQA} & \textbf{SocIQA} & \textbf{WinoGr.} & \textbf{Avg.}\\
      \midrule
      Indep-MoE & 55.18 & 56.27 & 48.19 & 40.73 & 35.80 & 66.70 & 39.46 & 55.96 & 49.79 \\
      LowRank-MoE & 58.84 & 59.42 & 48.26 & 40.68 & 33.00 & 66.54 & 40.69 & 54.38 & 50.22 \\
      PathMoE & 57.49 & 60.06 & 48.30 & 40.69 & 34.40 & 67.25 & 39.87 & 56.99 & \textbf{50.63} \\
      PathB8-MoE & 58.16 & 58.20 & 48.26 & 40.35 & 34.40 & 66.65 & 40.33 & 54.38 & 50.09 \\
      \bottomrule
  \end{tabular}
  }
\end{table}

\subsection{Orthogonal Benefits with Other Routing Methods}
\label{app:orthogonal}

Block-wise shared routing provides orthogonal benefits that can be combined with other routing improvements. Table \ref{tab:orthogonal} shows the performance of applying \pathmoe{} (block size 4) on top of X-MoE, which routes tokens on a low-dimensional normalized representation space. Load balancing losses are used for both models.

\begin{table}[h!]
  \centering
  \caption{Combining \pathmoe{} with X-MoE on Fineweb-100B (0.9B total / 0.37B active). PathMoE improves X-MoE's average accuracy (standard deviation is $\sim0.28\%$).}
  \label{tab:orthogonal}
  \resizebox{\columnwidth}{!}{%
  \begin{tabular}{lccccccccc}
      \toprule
      \textbf{Routing} & \textbf{ARC-E} & \textbf{BoolQ} & \textbf{HSwag} & \textbf{LAMB.} & \textbf{OBQA} & \textbf{PIQA} & \textbf{SocIQA} & \textbf{WinoGr.} & \textbf{Avg.} \\
      \midrule
      X-MoE & 43.27 & 60.55 & 46.26 & 45.22 & 32.20 & 67.63 & 39.61 & 52.80 & 48.44 \\
      PathB4X-MoE & 44.07 & 61.07 & 47.29 & 48.30 & 30.80 & 66.43 & 39.97 & 52.96 & \textbf{48.86} \\
      \bottomrule
  \end{tabular}
  }
\end{table}

The improvement demonstrates that cross-layer coordination through block-wise shared routing complements representation-based routing enhancements, suggesting that \pathmoe{} can be applied as a general technique on top of existing routing methods.

\subsection{Expert Path Concentration: Full Results}
\label{app:path_concentration}

Table \ref{tab:path_ablation_full} reports per-task results for the path restriction experiment described in Section~\ref{sec:path_conc}.

\begin{table}[h!]
  \centering
  \caption{Full per-task results for expert path restriction. Models are trained with routing restricted to the top-$K$ most frequent paths identified at 5k steps (1.25\% of training).}
  \label{tab:path_ablation_full}
  \resizebox{\columnwidth}{!}{
  \begin{tabular}{lcccccccccc}
      \toprule
      \textbf{\# Paths} & \textbf{ARC-E} & \textbf{BoolQ} & \textbf{HSwag} & \textbf{LAMBADA} & \textbf{OBQA} & \textbf{PIQA} & \textbf{SocIQA} & \textbf{WinoGr.} & \textbf{Avg.} \\
      \midrule
      10 & 40.45 & 59.54 & 39.34 & 40.87 & 30.40 & 66.81 & 37.82 & 54.54 & 46.22 \\
      100 & 43.81 & 56.79 & 47.04 & 45.10 & 30.80 & 67.46 & 40.23 & 54.22 & 48.18 \\
      500 & 44.74 & 59.42 & 48.24 & 47.58 & 31.20 & 66.00 & 39.97 & 53.75 & 48.86 \\
      All & 44.82 & 56.48 & 47.76 & 46.42 & 31.60 & 66.76 & 39.41 & 56.43 & 48.71 \\
      \bottomrule
  \end{tabular}
  }
\end{table}

\subsection{Token Category Definitions}

For our path specialization analysis, we classify tokens into detailed linguistic categories. Section \ref{tab:token_categories} provides the complete list of categories used in our analysis.

\begin{table}[h!]
  \centering
  \caption{Token categories used for path specialization analysis. Each category is defined by curated word lists or morphological patterns.}
  \label{tab:token_categories}
  \begin{tabular}{llp{7cm}}
      \toprule
      \textbf{Category} & \textbf{Type} & \textbf{Examples} \\
      \midrule
      \texttt{person\_names} & Lexical & Andrea, Richard, Oprah, Mary, John, Elizabeth \\
      \texttt{titles\_roles} & Lexical & secretary, minister, commander, winner, professor, CEO \\
      \texttt{speech\_verbs} & Lexical & said, explained, told, asked, claimed, announced \\
      \texttt{adverbs\_discourse} & Lexical & especially, actually, particularly, however, therefore \\
      \texttt{adverbs\_manner} & Lexical & quickly, carefully, successfully, directly, properly \\
      \texttt{adverbs\_time} & Lexical & now, today, recently, always, sometimes, currently \\
      \texttt{adverbs\_other} & Pattern & Words ending in -ly not in above lists \\
      \texttt{nationalities} & Lexical & American, British, Chinese, European, Japanese \\
      \texttt{temporal\_words} & Lexical & Monday, January, morning, year, summer, afternoon \\
      \texttt{prepositions} & Lexical & in, on, at, to, for, with, by, from, about \\
      \texttt{conjunctions} & Lexical & and, or, but, because, although, while, if \\
      \texttt{determiners} & Lexical & the, a, an, this, that, some, every, each \\
      \texttt{pronouns} & Lexical & he, she, they, it, who, someone, anybody \\
      \texttt{quantifiers} & Lexical & all, many, most, million, percent, several \\
      \texttt{common\_verbs} & Lexical & is, have, go, make, take, know, said, get \\
      \texttt{verbs} & Pattern & Words ending in -ing or -ed (running, created) \\
      \texttt{adjectives} & Lexical & good, new, important, political, economic, public \\
      \texttt{proper\_nouns} & Pattern & Capitalized words not in name list \\
      \texttt{abstract\_nouns} & Pattern & Words ending in -tion/-sion/-ness/-ment \\
      \texttt{agent\_nouns} & Pattern & Words ending in -er/-or (player, actor) \\
      \texttt{numbers} & Pattern & Numeric digits (1, 2, 100, 2024) \\
      \texttt{ordinals} & Pattern & Numeric ordinals (1st, 2nd, 3rd) \\
      \texttt{punctuation} & Pattern & Punctuation marks (. , ! ? ; : ' " ( )) \\
      \bottomrule
  \end{tabular}
\end{table}

\end{document}